# MACHINE LEARNING MODEL FOR PREDICTING SURFACE WETTABILITY IN LASER-TEXTURED METAL ALLOYS

Macro#: 339

Mohammad Mohammadzadeh Sanandaji[1], Danial Ebrahimzadeh[2], Mohammad Ikram Haider[1], Yaser Mike Banad[2], Aleksandar Poleksic[3] and Hongtao Ding*[1]

[1]Department of Mechanical Engineering, University of Iowa, Iowa City, IA 52242, USA
[2]School of Electrical and Computer Engineering, University of Oklahoma, USA
[3]Department of Computer Science, University of Northern Iowa, Cedar Falls, Iowa, USA
*Corresponding author. Tel: 319-335-5813; Email: hongtao-ding@uiowa.edu

**Abstract**

Surface wettability, governed by both topography and chemistry, plays a critical role in applications such as heat transfer, lubrication, microfluidics, and surface coatings. In this study, we present a machine learning (ML) framework capable of accurately predicting the wettability of laser-textured metal alloys using experimentally derived morphological and chemical features. Superhydrophilic and superhydrophobic surfaces were fabricated on AA6061 and AISI 4130 alloys via nanosecond laser texturing followed by chemical immersion treatments. Surface morphology was quantified using the Laws texture energy method and profilometry, while surface chemistry was characterized through X-ray photoelectron spectroscopy (XPS), extracting features such as functional group polarity, molecular volume, and peak area fraction. These features were used to train an ensemble neural network model incorporating residual connections, batch normalization, and dropout regularization. The model achieved high predictive accuracy ($R^2 = 0.942$, RMSE = 13.896), outperforming previous approaches. Feature importance analysis revealed that surface chemistry had the strongest influence on contact angle prediction, with topographical features also contributing significantly. This work demonstrates the potential of artificial intelligence to model and predict wetting behavior by capturing the complex interplay of surface characteristics, offering a data-driven pathway for designing tailored functional surfaces.

## 1. Introduction

Natural surfaces, such as lotus leaves and insect wings, often exhibit intricate nanostructures that give rise to remarkable wetting behaviors. Natural surfaces often possess fine nanostructures that lead to unique functional properties. Inspired by these biological systems, researchers have begun engineering nanostructured metal alloys to control surface wettability and enhance liquid–surface interactions in various applications. Wettability is commonly characterized by the contact angle formed at the interface between liquid, solid, and gas phases. Surfaces exhibiting extreme wetting behaviors such as superhydrophobic (contact angle > 150°) or superhydrophilic (contact angle < 10°) have attracted significant attention due to their practical advantages. For example, superhydrophobic surfaces are useful in anti-icing [1]self-cleaning [2], and drag reduction [3] applications, while superhydrophilic surfaces are essential in enhancing liquid spreading microfluidics, catalysis[4] , and biomedical integration [5,6].

The interaction between water and solid surfaces is governed by a complex combination of surface chemistry, topography, and environmental conditions. Traditionally, researchers have manipulated wettability by modifying both surface topography and surface chemistry using methods such as plasma techniques [7], wet chemical etching [8], lithography[9,10]and laser texturing [11]. These approaches allow surfaces to be engineered with tuneable wetting behaviors, ranging from superhydrophobicity to superhydrophilicity.

Building on these traditional methods, laser surface texturing has gained significant attention as an effective method to modify surface topography [12–14]. Ultrashort pulse lasers, such as femtosecond and picosecond lasers, are commonly employed to produce laser-induced periodic surface structures (LIPSS) [15,16] or to create hierarchical and dual-scale textures on metal surfaces [19,20]. For example, Granados et al. [17] found that femtosecond laser treatment forms laser-induced periodic surface structures (LIPSS) on boron-doped diamond, which makes the surface more hydrophilic. Femtosecond laser texturing of Ti-6Al-4V alloys has also been shown to produce hydrophilic surfaces with water contact angles ranging from about 24° to 76° immediately after treatment [18].

In addition to modifying surface topography, altering the surface chemical composition plays a crucial role in defining wettability. Recently, different chemical treatment methods have been used to induce superhydrophobic and superhydrophilic coatings onto laser-textured metal surfaces. Chemical treatments are often employed post-laser processing to functionalize the textured surfaces with specific molecular groups. For example, chemical modification using fluorosilane compounds is commonly performed following laser-induced surface structuring to induce superhydrophobic behavior on metal alloys. A single layer of functionalized fluorosilane was applied to the textured surface to lower its surface energy. Wu et al. [19] used chemical vapor deposition (CVD) to apply 97% trichloro (1H,1H,2H,2H-perfluorooctyl) silane [$CF_3(CF_2)_5(CH_2)_2SiCl_3$] onto laser-created microconical patterns on stainless steel surfaces, successfully inducing superhydrophobicity. Likewise, Yang et al. [20] reduced the surface energy of laser-ablated aluminum by immersing the samples in an ethanol-based solution containing (heptadecafluoro-1,1,2,2-tetradecyl) triethoxysilane for two hours, leading to a superhydrophobic surface. Following a comparable method, Rajab et al. [21] treated laser-textured stainless-steel samples by soaking them for two hours in a 1% methanol solution of heptadecafluoro-1,1,2,2-tetrahydrodecyl-1-trimethoxysilane [$CF_3(CF_2)_7(CH_2)_2Si(OCH_3)_3$], which resulted in stable superhydrophobicity. Pendurthi et al. [22] applied heptadecafluoro-1,1,2,2-tetrahydrodecyl trichlorosilane [$CF_3(CF_2)_7(CH_2)_2SiCl_3$] to titanium, aluminum, and various stainless steel substrates following laser texturing, resulting in superhydrophobic characteristics. Studies have shown that the synergistic combination of surface topography and surface chemistry is essential to achieve extreme wetting states, such as those exhibiting both high contact angle and low hysteresis.

Despite substantial progress in experimental techniques, predicting wettability remains a formidable challenge. Recent advances in artificial intelligence and machine learning have opened new avenues for predicting complex material properties, including surface wettability [23]. Machine learning models offer the capability to capture nonlinear relationships between multiple surface parameters and wettability outcomes, overcoming the limitations of classical analytical models [24]. Several pioneering studies have demonstrated the potential of ML approaches in surface science: Huang et al. [25] successfully employed XGBoost to predict contact angles by considering both surface topography and chemistry parameters, achieving significant improvements over traditional methods. Rabbani et al. [26] applied convolutional neural networks to analyze microscopy images of textured surfaces, directly predicting contact angles from surface morphology with remarkable accuracy. Furthermore, Wei et al. [27] developed a hybrid ML approach combining random forests and support vector machines to predict dynamic wetting behavior, highlighting the versatility of ensemble methods in capturing complex surface phenomena.

Despite these advances, existing ML models for wettability prediction have notable limitations: they often focus on either topography or chemistry in isolation, utilize relatively simple architectures, or achieve limited prediction accuracy for complex laser-textured surfaces. In this work, we present a breakthrough neural network ensemble model that addresses these limitations through several key innovations. Our model integrates comprehensive surface chemistry features (including functional group polarities, volumes, and area fractions) with topographical parameters in a sophisticated architecture employing residual connections, batch normalization, and advanced ensemble learning techniques. This approach achieves unprecedented prediction accuracy ($R^2$ = 0.942, RMSE = 13.896) for laser-textured metal surfaces which is a significant improvement over previous ML methods which typically report $R^2$ values below 0.95 [25,26].

The primary contributions of this work are threefold: (1) We develop the first neural network ensemble model that simultaneously considers both nanoscale chemical functional groups ($CF_2$, $CF_3$, CN) and microscale topographical features for wettability prediction, providing a holistic understanding of surface-liquid interactions; (2) We introduce advanced architectural innovations including residual connections and multi-model ensembling that enable the capture of highly nonlinear relationships between surface parameters and contact angles; and (3) We validate the dominant role of surface chemistry in determining wettability through rigorous feature importance analysis, providing critical insights for the design of surfaces with targeted wetting properties. Beyond achieving state-of-the-art accuracy, our model provides interpretable insights that can directly inform the design of next-generation functional surfaces.

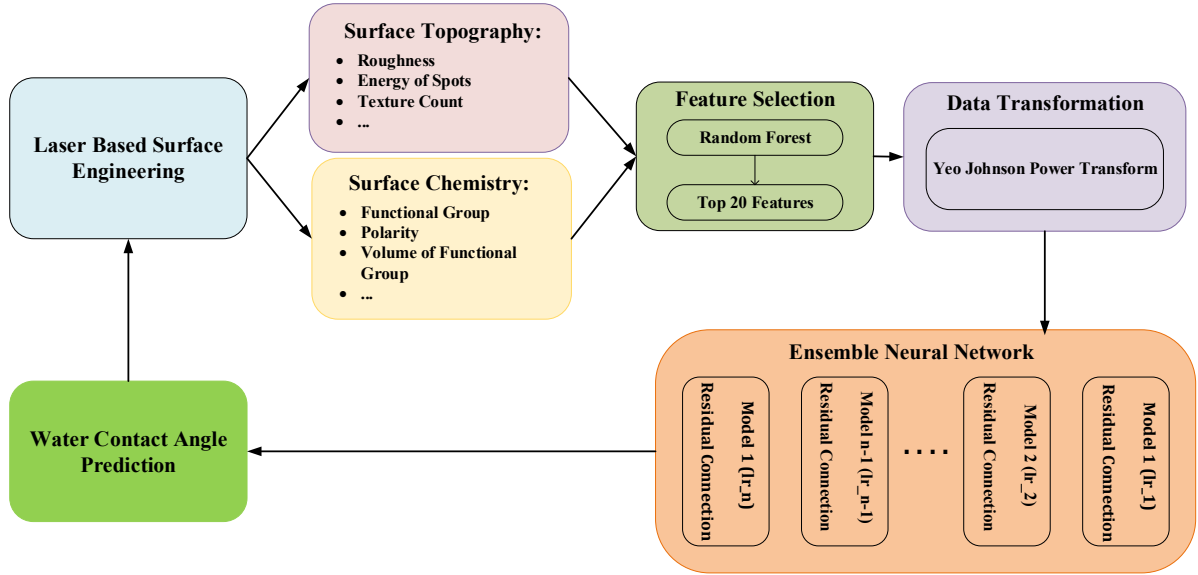

Fig1. General framework of machine learning of surface wetting behaviors

## 2. Laser Experiments & Input Data Collection

In this work, a novel nanosecond laser-based high-throughput surface nanostructuring (nHSN) process was employed, consisting of two sequential steps: (1) laser surface texturing and (2) chemical immersion treatment (CIT). In the first step, nanosecond laser texturing was conducted using a Q-switched Nd:YAG laser operating at a wavelength of 1064 nm, with pulse durations ranging from 8 to 12 ns and pulse energies in the hundreds of millijoules. Beam delivery and surface scanning were carried out using a 3-axis galvanometric scanner (SCANLAB intelliSCAN® 20 paired with a varioSCANde 40i) equipped with an f-theta lens.

As illustrated in Fig. 2, the laser texturing process involved raster scanning the entire substrate surface while it was submerged in deionized (DI) water. The presence of DI water during laser processing helps confine the laser-induced plasma and enhance surface modification effects. In this study, two common metal alloys, aluminum alloy 6061 (AA6061) and steel alloy AISI 4130, were used as substrate materials. A total of 60 AA6061/AISI 4130 samples were treated using varying laser power intensities, ranging from 0.1 $GW/cm^2$ to 8.4 $GW/cm^2$, while keeping other laser parameters, such as beam diameter and pulse width, constant. The objective was to systematically vary the surface topography for further analysis and data collection.

In the next step, to modify surface chemistry, samples were immersed in a chemical solution to attach desired chemical functional groups on the laser treated surfaces. For the superhydrophilic treatment, samples were immersed in an ethanol solution containing 1.5 vol% of 3-cyanopropyltrichlorosilane (CPTS) at room temperature for approximately 3 hours, after treatment, the samples were rinsed with DI water, dried with compressed air, and then placed in a vacuum oven at 80 °C for 1 hour to ensure complete drying. For the superhydrophobic samples, a similar procedure was followed, except the workpieces were immersed in an ethanol solution of 1.5 wt% containing one of three different silane-based reagents: 1H,1H,2H,2H-perfluorooctyltrichlorosilane (FOTS), 1H,1H,2H,2H-perfluorodecyltrichlorosilane (FDTS), and 1H,1H,2H,2H-perfluorododecyltrichlorosilane (FDDTS).

After sample fabrication, the static water contact angle (θw) of the treated surfaces was measured using a contact angle goniometer (Rame-Hart Model 100) equipped with a high-resolution CMOS camera (Thor Laboratories, 6–60× magnification). For each measurement, a 5 μL droplet of deionized water was gently dispensed onto the surface, and the droplet profile was recorded for analysis. To ensure measurement reliability and account for surface heterogeneity, five measurements were taken at different locations on each specimen. The average of these values was computed and used as the output data for the machine learning model.

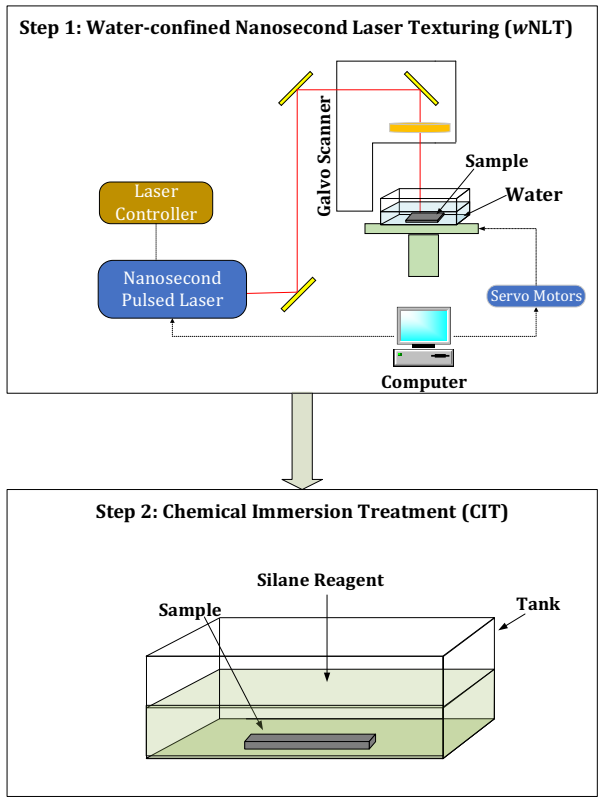

Fig. 2. Experimental set up (1) nanosecond laser texturing (NLT) and (2) chemical immersion treatment (CIT)

## 2.1 Surface Topography Data Collection

We employed a multi-scale approach to characterize the surface topography of the nHSN specimens by extracting both microscale and nanoscale features. The microscale topography was quantified by measuring the arithmetical mean roughness $S_a$ with a KEYENCE VR profilometer. These $S_a$ values served as a direct input for the machine learning model. For domains requiring higher resolution, a distinct analysis of nanoscale surface features was performed and listed in Table -1. We introduced a novel application of the laws texture energy method for the quantitative analysis of nanoscale surface features on laser-textured alloys. Standard microscale characterization methods lack the necessary resolution to capture these critical nanoscale topographies, necessitating the innovative approach presented herein. By focusing on the resulting energy maps, our adapted protocol enables reliable quantification of nanoscale surface features.

Fig. 3 describes the roughness data of AA6061 samples fabricated with nHSN process with hydrophobic functionalization. Compared to laser textured surfaces fabricated in air which produce deep microgrooves having depth on the order several hundred microns, the nHSN process produces nanoscale features on surface. From Fig. 3b the roughness values for super hydrophobic AA6061 samples varies from 500 nm to 1050 nm. Similar range of values were obtained for super hydrophilic AA6061 samples too. However, for AISI 4130 steel samples the roughness value varied from 600 nm to 1150 nm for both hydrophobic and hydrophilic cases. From Fig. 3b, it is evident that roughness values increase with increasing power intensities. However, there is a sharp increase in roughness values when power intensities values increase from 0.2 to 1.2 GW/cm$^2$. After that the increase in roughness values is minimal with respect to increasing laser power intensities.

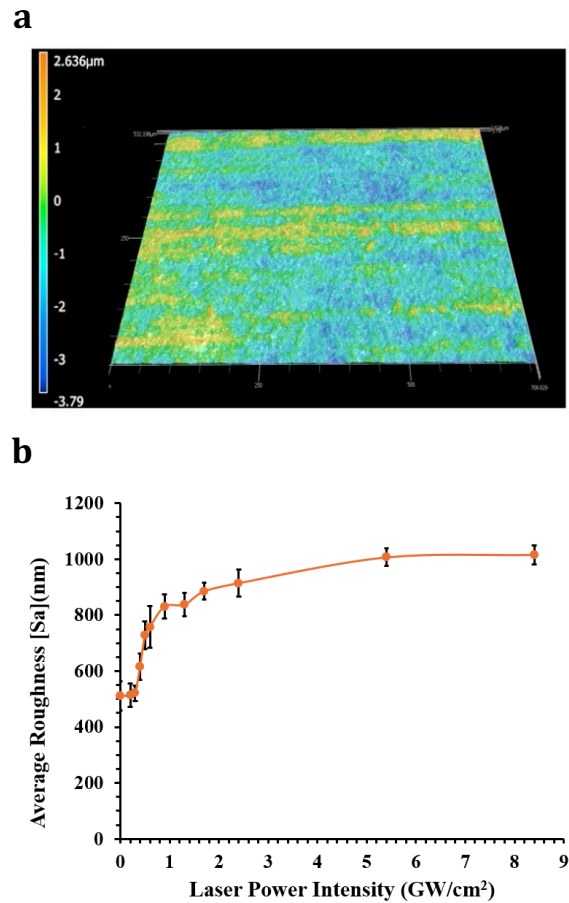

Fig. 3. Microscale surface roughness data over a domain size of 640x480 um.: (a) typical 3D surface profilometry over 640x480 um; (b) relation between the laser processing parameter (in this case laser power intensity) and surface roughness for superhydrophobic nHSN AA6061 sample with different power intensities.

Table 1. Parameters of surface topography as ML mode input

| Length scale | Parameter | Measurement | Definition | Algorithm |
|---|---|---|---|---|
| Microscale (Hundreds to tens of μm) | $S_a$ | KEYENCE VR profilometer | Average of the absolute deviations of the surface profile from its mean line | Statistical Measurements |
| Submicron & Nanoscale | Feature Area, $A_n$ | SEM | Average Area of each nanoscale feature (Ripples, Spots, Waves, Level, Edge). | Laws method |
| | Texture Count, $T_n$ | SEM | The Texture Count of pixels in the image that has been selected as nanoscale features. | Laws method |
| | Feature Energy, $E_n$ | SEM | Average Energy of Different Laws Feature | Laws method |

The nanoscale texture analysis was specifically focused on identifying characteristic features including edges, spots, ripples, waves, and levels, which are significant for understanding surface wettability transitions from hydrophilic to hydrophobic states induced by varied laser powers and subsequent chemical treatments. SEM images of both AA6061 and AISI 4130 fabricated with different power intensities and different post processing condition was analyzed to calculate the feature values. nHSN images were kept at high resolution (2560 X 1920) and high magnification (50000X) to maintain the consistency of data to put as an input of our ML model.

Initially, SEM images were converted into grayscale format. Grayscale conversion simplifies computational complexity by eliminating color information and focusing solely on intensity variations crucial for texture analysis [28][29]. Then laws texture energy method was applied to the greyscale images to compute different feature values. Kenneth I. Laws invented the texture energy method introducing five 1D masks (L5, E5, S5, R5, W5) and their 2D separable 5×5 kernels to extract local texture energy via absolute-value filtering and moving window averaging. Laws distilled this into a real-time algorithm: convolve an image with a subset of the 5×5 filters, take the absolute response to form energy maps, optionally smooth, and then apply simple thresholding or per-pixel classification to segment textured scenes [30,31]. Modern enhancements include multi-resolution laws' masks, which fuse dyadic wavelet transforms with the classic 5×5 Laws kernels to improve texture classification on benchmarks[32] and DWT-based laws masks applied to SEM images for automated defect detection in quality inspection, greatly boosting industrial classification performance. Parallelly, raw texture energy statistics (mean, standard deviation, entropy) have been fed into k-NN, SVM, and ELM classifiers to achieve tremendous accuracy in species recognition, showcasing Laws' method's versatility in modern machine vision tasks [33,34]. Laser texturing generates different surface structures whose spatial frequencies match exactly the ripple, edge and wave filters in the mask bank, so laws' method provides direct, quantitative metrics (count, mean energy, area) that correlate with processing parameters and resulting functional properties (e.g., wettability). This approach is novel in the field of laser texturing as it offers a more robust and automatable alternative to traditional methods like profilometry or FFT, which may not effectively differentiate co-existing local features.

Each grayscale image underwent convolution using laws' texture kernels, producing specific energy maps defined by the absolute magnitude of the filtered images. These kernels effectively detect microtextural variations by highlighting specific texture attributes, thereby providing quantifiable texture energy distributions across the surface [31,35]. Each kernel emphasizes a different type of feature (Table-2).: the 'Edge' kernel detects abrupt intensity transitions, corresponding to sharp boundaries or linear defects; the 'Spot' kernel highlights isolated features, such as pits or inclusion[36]. Level (L5×L5) mask captures coarse brightness and is often used just to normalize contrast. The Edge mask (E5×E5) highlights abrupt linear transitions – e.g. scratches or fracture lines. The Ripple (R5×R5) and Wave (W5 ×W5) masks capture more periodic or two-dimensional undulations – e.g. fine toolmark ripples, dimple patterns, or repetitive etched patterns on the surface[37]. Quantities like texture count or mean area typically refer to the number and size of connected regions above a threshold in each filtered image, while mean of energy refer to the average intensity or variation of the laws-filtered image. Together, these energy-based descriptors can effectively quantify nanoscale texture on metal surfaces, providing features that correlate with material condition or processing history [36,37]. If $F_n[i,j]$ is the result of filtering the preprocessed image after n-th mask, at pixel [i, j], then texture energy map $E_n[r,c]$ is computed by [38]:

$$E_n(r, c) = \sum_{j=c-7}^{c+7} \sum_{i=r-7}^{r+7} | F_n[i,j] | \quad (1)$$

Table 2. Texture energy measurement vectors for this ML study

| Name of Features | Vector Lengths | 2D Mask Vectors | Description of 2D mask |
|---|---|---|---|
| Level | L5 = [1 4 6 4 1] | L5 X L5 | Level detection in horizontal and vertical direction of 5 neighboring pixels |
| Edge | E5 = [-1 -2 0 3 1] | E5 X E5 | Edge detection in horizontal and vertical direction of 5 neighboring pixels |
| Spot | S5 = [-1 0 2 0 -1] | S5 X S5 | Spot detection in horizontal and vertical direction of 5 neighboring pixels |
| Wave | W5 = [-1 2 0 -2 1] | W5 X W5 | Wave detection in horizontal and vertical direction of 5 neighboring pixels |
| Ripple | R5 = [1 -4 6 -4 1] | R5 X R5 | Ripple detection in horizontal and vertical direction of 5 neighboring pixels |

To ensure consistent and unbiased segmentation of relevant texture features, Otsu's thresholding method was applied to the generated energy maps. This automatic global thresholding approach effectively discriminates regions of distinct energy levels by minimizing intra-class variance and maximizing inter-class variance, enhancing segmentation reliability [39].

Following threshold determination, binary segmentation was conducted, transforming energy maps into binary images where pixels exceeding the threshold were labeled distinctly. This critical step isolates prominent texture features pertinent to the assessment of surface wettability modifications. After that law's energy, texture count and area of each feature (e.g. Spots, Waves, Edges, Ripples, Levels) were calculated[36] to generate nanoscale and sub-micron level data.

Fig. 4 shows comparison between untreated AA6061 and nHSN treated AA6061 samples. While visually it can be seen that nHSN sample is more chaotic and has random structures than untreated one (Fig. 4 a-b), laws texture energy method quantifies the difference with numerical values of texture count and energy value of edges, spots, ripples and waves. The magnification of the SEM images was 5000X and field of view was set to 29 μm x 22 μm. To cancel the noise from the images several post processing algorithms like thresholding, kernel size increasing etc. were considered. From Fig. 4 c, it can be seen that nHSN samples can create nanoscale features as the values of texture count is approximately 20-25 times higher than untreated samples.

We implemented a multi-scale characterization framework that integrates conventional microscale profilometry with a novel application of the laws texture energy method for nanoscale feature analysis from SEM images. The novelty of employing texture energy analysis on laser-processed alloys provides a robust and automatable alternative to traditional characterization techniques. Ultimately, this process yields quantitative, high-fidelity descriptors of the surface topology that are directly correlated with processing parameters and functional properties, thereby establishing them as highly suitable inputs for predictive machine learning models.

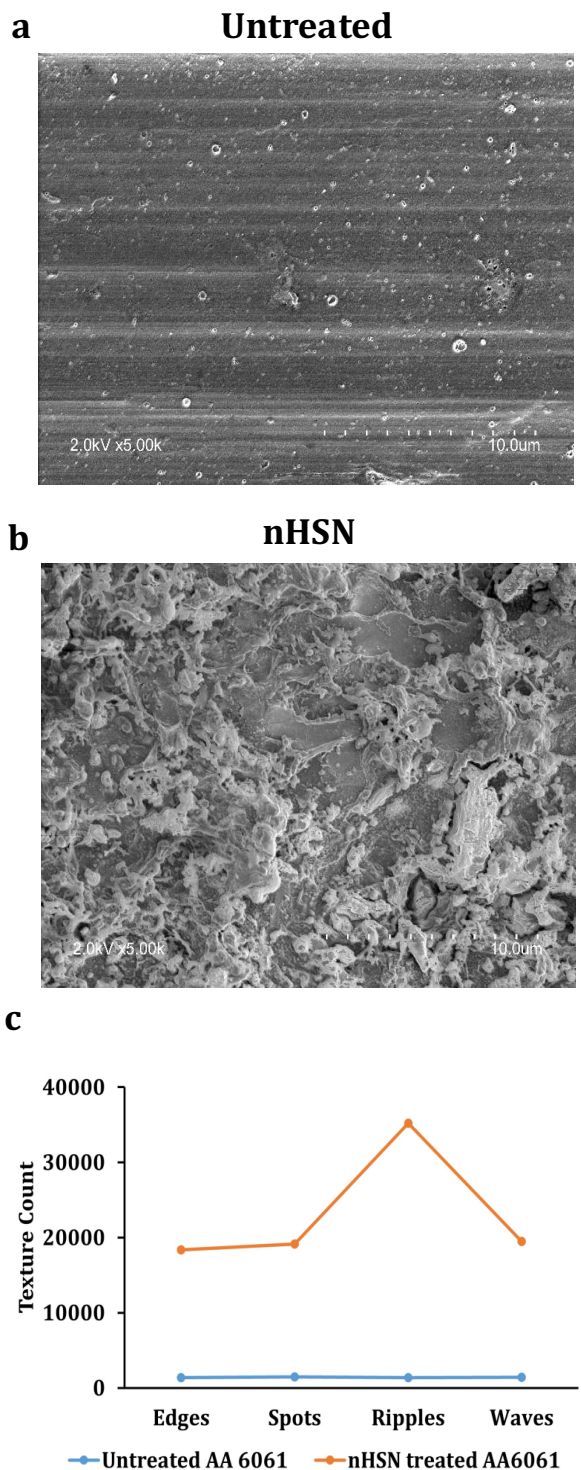

Fig. 4. Nanoscale feature analysis: (a) SEM images of untreated raw sample surface; (b) SEM images of nHSN treated surface; (c) comparison of texture count ($T_n$) for untreated and treated surfaces.

## 2.2 Surface Chemistry Data Collection

The surface chemistry of the nHSN samples plays a crucial role in determining its wettability characteristics. To analyze this aspect, X-ray photoelectron spectroscopy (XPS) was employed using a Kratos Axis Ultra high-performance system. In the case of the superhydrophobic samples, those subjected to both the wNLT and subsequent CIT processes a notably different XPS spectrum was observed. In addition to the typical elements such as oxygen, carbon, and aluminum two new peaks corresponding to fluorine and silicon appeared in the survey spectrum (Fig. 5a). These elements originated from the FOTS [$CF_3(CF_2)_5(CH_2)_2SiCl_3$; 98%] reagent. Further analysis of the carbon core-level spectrum revealed a combination of -$CH_2$-, -$CF_2$-, and -$CF_3$- groups, which are characteristic of the FOTS molecular structure (Fig. 5b). Samples treated with FDTS [$CF_3(CF_2)_7(CH_2)_2SiCl_3$; 97%] and FDDTS [$CF_3(CF_2)_9(CH_2)_2SiCl_3$; 97%] are expected to show similar spectral features, as these compounds share the same functional groups found in FOTS. The primary distinction among the three fluorosilane reagents (FOTS, FDTS, and FDDTS) lies in the length of the $-CF_2-$ chains within their functional groups. Specifically, the number of $-CF_2-$ units is five in FOTS, seven in FDTS, and nine in FDDTS. Avik et al. demonstrated that an increase in the number of $-CF_2-$ groups on the treated surface correlates with a higher contact angle (θ), indicating enhanced hydrophobicity [40].

For the superhydrophilic samples, the surface chemistry of the nHSN material treated with CPTS [$CN(CH_2)_3SiCl_3$; 97%] was analyzed. As shown in Fig. 5c, the XPS spectrum of the nHSN surface modified via the CIT process with CPTS exhibits notable differences compared to the surface treated with FOTS. Specifically, the survey spectrum of the highly hydrophilic nHSN surface revealed two additional peaks corresponding to nitrogen and silicon, alongside the typical signals for oxygen, carbon, and aluminum. These nitrogen and silicon elements originate from the CPTS reagent, which contains two key functional groups: $-CN$ and $-CH_2-$ (Fig. 5d). These groups bond to the laser-textured surface during treatment, introducing nitrile functionalities. The $-CN$ group is highly polar and plays a key role in enhancing the hydrophilicity of the surface. Unlike some prior studies that used XPS binding energies as machine learning inputs [41], we found no direct correlation between binding energy and wettability. Binding energy primarily serves to identify chemical species, not to predict contact angle values. Instead, we focused on features that directly reflect surface composition and interaction potential. As a result, three surface chemistry descriptors were selected as input features for the ML model:

(1) **Peak Area Fraction ($P_{AF}$):** This represents the relative abundance of each functional group, calculated by dividing its spectral peak area by the total peak area of related elements. It captures surface coverage of specific chemistries.

(2) **Dipole Moment (μ):** The polarity of the most dominant surface group, which influences how strongly the surface interacts with water molecules.

(3) **Molecular Volume ($V_g$):** The physical size of the dominant group, which affects packing density and the extent of surface-liquid interactions.

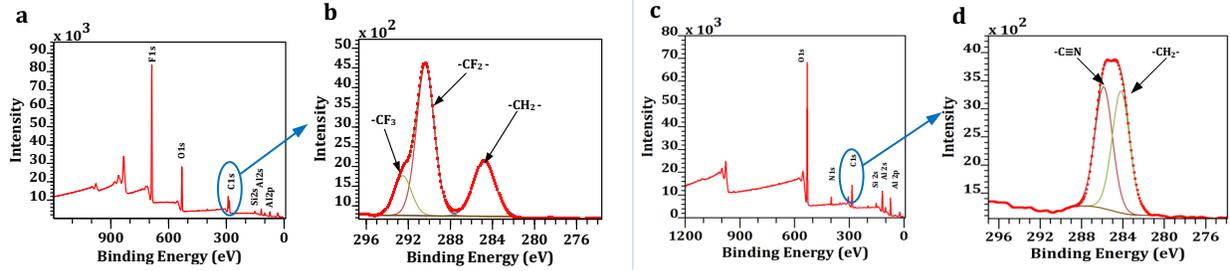

Fig. 5. XPS survey spectra of the surface layer for superhydrophobic and superhydrophilic AA6061: (a) nHSN surface chemically treated with FOTS reagent; (b) Core-level analysis of the carbon element; (c) nHSN surface chemically treated with CPTS reagent; (d) Core level XPS analysis for carbon and nitrogen element.

The most dominant groups were identified as –$CF_2$– for fluorosilane-treated (hydrophobic) surfaces and –CN for CPTS-treated (hydrophilic) ones. Together, these parameters quantify the chemical makeup and physicochemical behavior of the surface, enabling accurate wettability prediction in the machine learning model.

## 3. Neural Network Ensemble Model

This study develops an advanced neural network ensemble model to predict contact angles of laser-textured metal surfaces based on their different features such as chemical and topographical characteristics. The proposed methodology integrates feature selection, data preprocessing, and a novel ensemble neural network architecture with residual connections to capture the complex relationships between surface properties and wettability behavior.

### 3.1 Feature Selection and Importance Analysis

The high-dimensional nature of surface characterization data presents significant challenges for machine learning models. To address this issue, we employ a Random Forest-based feature selection strategy [42]. Random Forest algorithms have demonstrated superior performance in identifying relevant features in materials science applications [43]. The algorithm constructs multiple decision trees and evaluates feature importance based on the mean decrease in impurity:

$$I_j = \frac{1}{N_T} \sum_{t=1}^{N_T} \sum_{n \in N_t^j} p(n)\Delta_n \qquad (2)$$

where $I_j$ represents the importance of feature $j$, $N_T$ is the number of trees in the forest, $N_t^j$ denotes the set of nodes where feature $j$ is used for splitting in tree $t$, $p(n)$ is the proportion of samples reaching node $n$, and $\Delta_n$ is the impurity decrease at node $n$. This approach systematically identifies the most informative surface parameters while reducing computational complexity and mitigating the curse of dimensionality.

### 3.2 Data Preprocessing and Transformation

Surface chemistry and topography measurements typically exhibit non-normal distributions with significant skewness and varying scales, which can adversely affect neural network convergence and performance. To address these challenges, we apply the Yeo-Johnson power transformation [44], which generalizes the Box-Cox transformation to handle both positive and negative values:

$$y_i^{(\lambda)} = \begin{cases} [(y_i + 1)^\lambda - 1]/\lambda & \text{if } \lambda \neq 0 \text{ and } y_i \geq 0 \\ \log(y_i + 1) & \text{if } \lambda = 0 \text{ and } y_i \geq 0 \\ -[(-y_i + 1)^{2-\lambda} - 1]/(2 - \lambda) & \text{if } \lambda \neq 2 \text{ and } y_i < 0 \\ -\log(-y_i + 1) & \text{if } \lambda = 2 \text{ and } y_i < 0 \end{cases}$$

(3)

The transformation parameter $\lambda$ is optimized using maximum likelihood estimation to achieve approximately normal distributions. This preprocessing step ensures that the neural network receives well-conditioned inputs, facilitating more effective gradient-based optimization.

### 3.3 Neural Network Architecture with Residual Connections

The core of our methodology is an advanced neural network architecture that incorporates residual connections to enable the training of deeper networks while avoiding gradient vanishing problems [45]. As shown in Fig. 6 the architecture begins with an input layer that transforms the selected features into a high-dimensional representation space. Subsequently, multiple hidden layers process this representation through a series of nonlinear transformations.

The key innovation in our architecture is the implementation of residual connections between layers with compatible dimensions. For each hidden layer $l$, the forward propagation follows:

$$h_{l+1} = f(W_l h_l + b_l) + h_{l-k} \quad (4)$$

where $f$ represents the activation function, $W_l$ and $b_l$ are the learnable weight matrix and bias vector, and $h_{l-k}$ is the residual connection from a previous layer with matching dimensions. We employ Leaky Rectified Linear Units (Leaky ReLU) as activation functions to mitigate the dying ReLU problem while maintaining computational efficiency [46].

Each hidden layer incorporates batch normalization [47] to address internal covariate shift and accelerate training convergence. This technique normalizes the inputs to each layer, maintaining stable distributions throughout the network depth. Additionally, dropout regularization is applied to prevent overfitting by randomly deactivating neurons during training [48].

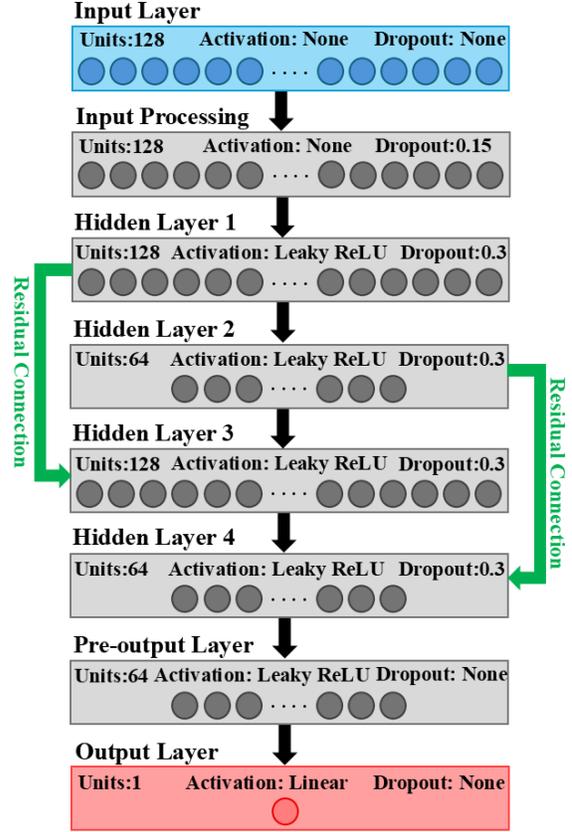

Fig. 6. Detailed architecture of the individual neural network used in the ensemble model.

### 3.4 Ensemble Learning Strategy

To enhance prediction robustness and reduce model variance, we implement an ensemble learning approach that combines multiple neural networks [49]. Ensemble methods have proven particularly effective in materials property prediction tasks [50]. Our strategy creates diversity among ensemble members through three mechanisms:

First, each model in the ensemble is trained with slightly different hyperparameters, particularly varying learning rates that encourage different convergence behaviors. Second, we employ adaptive learning rate scheduling with different patience parameters for each model, allowing them to explore different regions of the parameter space. Third, we utilize a composite loss function that combines Mean Squared Error (MSE) with Huber loss [51] to balance sensitivity to outliers with robust estimation:

$$\mathcal{L} = \alpha \mathcal{L}_{MSE} + (1 - \alpha)\mathcal{L}_{Huber} \quad (5)$$

The Huber loss provides robustness against outliers while maintaining differentiability, making it particularly suitable for regression tasks with potentially noisy measurements. The final ensemble prediction is obtained through model averaging, which has been shown to reduce prediction variance and improve generalization [52].

*3.5 Optimization Strategy*

The training process employs the AdamW optimizer [53], which decouples weight decay from gradient-based optimization, providing better generalization performance compared to traditional L2 regularization. We implement gradient clipping to prevent exploding gradients, a common issue in deep neural networks. Early stopping based on validation performance prevents overfitting while ensuring optimal model capacity utilization.

The proposed methodology represents a significant advancement in machine learning approaches for surface property prediction. By combining sophisticated preprocessing, advanced neural architecture design, and ensemble learning, our approach captures the complex nonlinear relationships between surface characteristics and wettability behavior. The residual connections enable effective training of deeper networks, while the ensemble strategy provides robust predictions across diverse surface conditions.

## 4. Results and Discussion

*4.1 Feature Selection and Feature Importance Ranking*

The Random Forest algorithm identified 20 key features from the initial 36-dimensional feature space that collectively explain the variance in water contact angle measurements. The selected features are categorized, showing that 85% (17 features) are chemical and 15% (3 features) are topographical. This distribution underscores the dominant role of surface chemistry in determining wettability, while also highlighting the complementary influence of topographical characteristics. Notably, topographical features such as surface roughness (Roughness nm) and texture descriptors (Spots S5S5 ME, Edges E5E5 ME) exhibit minimal correlation ($|r| < 0.3$) with chemical features, indicating their orthogonal contribution to the model. This independence supports the notion that wettability arises from the interplay of chemical composition and physical structure [54,55]. The Random Forest algorithm's retention of both highly correlated chemical features and independent topographical parameters suggests that even seemingly redundant chemical descriptors may capture subtle, yet important, variations relevant to accurate contact angle prediction.

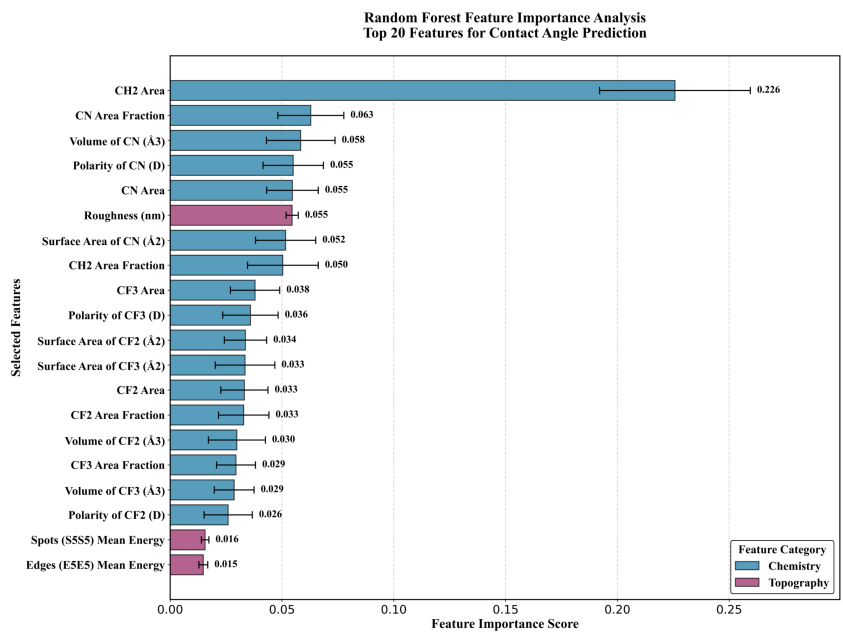

Fig. 7. Random Forest feature importance analysis for contact angle prediction. It is noted that results are restricted to the nHSN process.

In addition, we evaluated the relative importance of each selected feature in predicting water contact angles. Fig. 7 displays the Random Forest feature importance scores for the 20 selected features, averaged over 10 independent runs, with error bars representing the standard deviation across iterations. While feature importance measures how much each feature contributes to the model accuracy, it should be noted that feature importance does not directly correlate with the target variable (in our case the contact angle).

The analysis reveals a clear hierarchical structure in the predictive power of surface parameters. Among all features, $CH_2$ Area emerges as the most influential, with an importance score of 0.226 approximately four times greater than the median feature importance (0.055). This prominence of hydrocarbon coverage aligns with established surface energy principles, where $CH_2$ groups exhibit intermediate wettability between highly hydrophobic fluorinated groups and strongly hydrophilic cyano groups.

The top five features $CH_2$ Area, CN Area Fraction, Volume of CN, Polarity of CN, and CN Area collectively account for 45.7% of the total feature importance, with four of these being cyano-related. This concentration underscores the strong hydrophilic influence of the cyano functional group, attributed to its highly polar C≡N bond [56]

Notably, surface roughness (importance = 0.055) ranks as the most significant topographical feature and sixth overall, reinforcing theoretical models such as the Wenzel and Cassie-Baxter frameworks, which posit that topography modulates but does not dominate wettability [57]. Meanwhile, fluorinated groups ($CF_2$ and $CF_3$) exhibit moderate importance (cumulative = 0.257), which may reflect the relatively low variance among fluorosilane-treated samples in the dataset.

### 4.2 Model Performance and Validation

To evaluate model performance, we employed two primary metrics: Root Mean Square Error (RMSE) and the coefficient of determination ($R^2$). RMSE provides insight into prediction accuracy by measuring the average magnitude of prediction errors, while $R^2$ indicates how well the model explains variance in the data. Fig. 8 presents a comprehensive comparison of model performance across different algorithms. Our ensemble neural network achieved a mean RMSE of 13.896 ± 4.245 and $R^2$ of 0.937 ± 0.042 using 8-fold cross validation repeated twice (16 total folds), representing a 22.6% improvement over the XGBoost model reported by Huang et al.[25] (RMSE = 17.94°). The ensemble approach also outperformed single neural network (RMSE = 14.192 ± 5.919 and $R^2$ = 0.935 ± 0.053) and random forest models (RMSE = 15.153 ± 7.524 and $R^2$ = 0.923 ± 0.07), validating the effectiveness of ensemble averaging in reducing prediction variance. The cross-validation analysis demonstrates the neural network ensemble's robust predictive performance across different data partitions. The model achieved consistent accuracy with RMSE values ranging from 8.567° to 21.813° and $R^2$ values spanning 0.859 to 0.984 across all folds. While occasional outlier folds reaching higher RMSE values suggest the presence of challenging samples in the chemical-topographical parameter space, the overall stability of the cross-validation results validates the model's reliability for deployment in practical surface engineering applications.

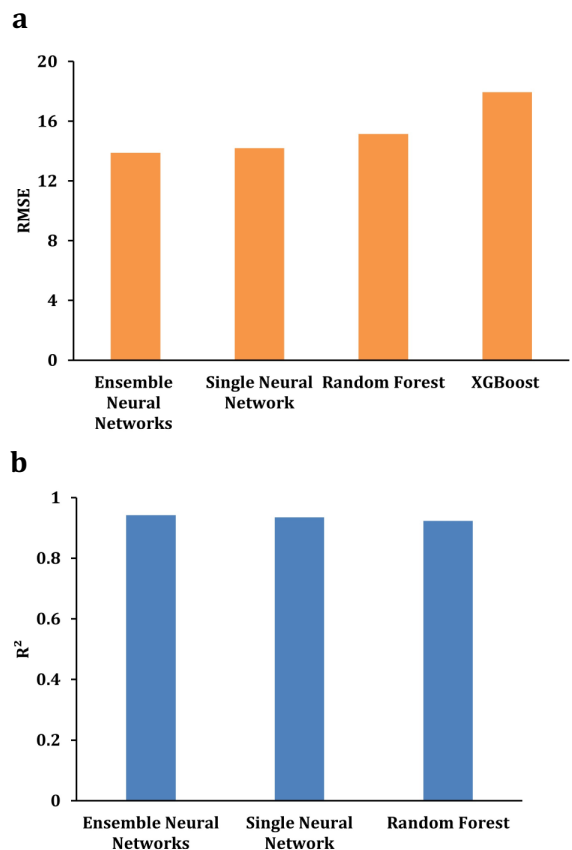

Fig. 8. comparison of model performance across different algorithms according to: (a) RMSE; (b) $R^2$

Fig. 9 presents a comprehensive validation of the neural network ensemble model through a predicted versus actual contact angle scatter plot, where each data point represents an individual prediction from the cross-validation analysis. The x-axis displays the experimentally measured contact angles (actual values), while the y-axis shows the corresponding neural network predictions across all cross-validation folds. The red dashed diagonal line (y=x) represents the theoretical line of perfect prediction, where predicted values would exactly equal measured values. The orange solid line represents the best-fit linear regression through the actual data points, with its equation displayed to quantify the relationship between predictions and measurements.

The tight clustering of data points around the perfect prediction line, combined with the near-overlap of the best-fit line with the diagonal reference, demonstrates exceptional model accuracy across the entire range of contact angles studied (from hydrophilic to superhydrophobic surfaces). The minimal scatter around the diagonal line indicates low prediction variance, while the proximity of the orange line to the red reference line confirms the absence of systematic prediction bias. This visualization validates that the neural network ensemble successfully captures the complex, non-linear relationships between the 20 selected surface features and contact angle behavior, demonstrating robust predictive performance suitable for practical surface engineering applications. The strong linear correlation between predicted and actual values across diverse surface conditions confirms the model's ability to generalize beyond the training data.

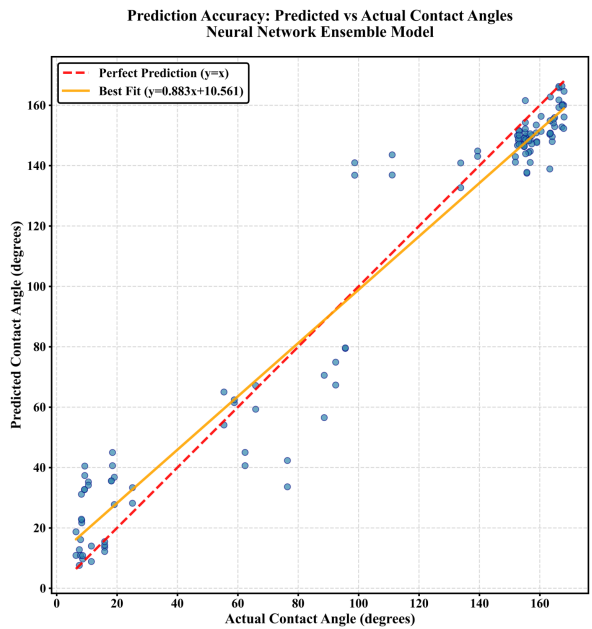

Fig. 9. Scatter plot of predicted versus actual water contact angles for the neural network ensemble model.

## Conclusions

In this work, we demonstrated that a machine learning model can reliably predict surface wettability based on detailed chemical and topographical data from laser-textured metal alloys. By combining SEM-based texture analysis with XPS-derived chemical features, the ensemble neural network achieved strong predictive performance ($R^2$ = 0.942, RMSE = 13.896). The results confirm that surface chemistry particularly polar functional groups like CN plays a dominant role in controlling contact angle, while surface roughness and texture provide complementary predictive power. Importantly, the model offers interpretable outputs that clarify the relationship between surface modification strategies and resulting wetting behavior. This approach provides a scalable and data-driven tool for designing functional surfaces with tailored wettability, especially in applications where precise wetting control is essential.

## Acknowledgement

The authors gratefully acknowledge the financial support from the National Science Foundation under Grant Number 2242763.

## Contribution

Mohammad Mohammadzadeh Sanandaji: Writing – original draft, Investigation, Formal analysis, Data curation, Conceptualization. Danial Ebrahimzadeh: Writing – original draft, Investigation, Formal analysis, Data curation, Conceptualization. Mohammad Ikram Haider: Writing, Investigation and formal analysis. Yaser Mike Banad: Supervision – review & editing. Aleksandar Poleksic: review & editing of data science analysis. Hongtao Ding: Writing – review & editing, Writing – original draft, Supervision, Resources, Funding acquisition, Formal analysis, Conceptualization.

## Declaration of Competing Interest

The authors declare that they have no known competing financial interests or personal relationships that could have appeared to influence the work reported in this paper.